\documentclass[twoside,11pt]{article}
\usepackage{jmlr2e}

\usepackage[numbers]{natbib}

\usepackage[utf8]{inputenc} 
\usepackage[T1]{fontenc}
\usepackage{url}
\usepackage{hyperref}       
\usepackage{booktabs}       
\usepackage{amsfonts}       
\usepackage{nicefrac}       
\usepackage{microtype}      
\usepackage{ifthen}
\usepackage[linesnumbered,ruled,algo2e]{algorithm2e}
\usepackage{amsthm, amssymb, amsmath}
\usepackage{graphicx}
\usepackage{cleveref}
\usepackage[colorinlistoftodos,prependcaption]{todonotes}
\usepackage{mathtools}

\usepackage{epstopdf}
\usepackage{algorithm}
 \usepackage{algorithmic}
\usepackage{wrapfig}
\usepackage{subcaption}
\usepackage{bbm}

\theoremstyle{plain}

\theoremstyle{definition}

\theoremstyle{remark}

\newcommand{\E}{\mathrm{E}}

\newcommand{\bbm}{\begin{bmatrix}}
\newcommand{\ebm}{\end{bmatrix}}

\newcommand{\R}{\mathrm{R}}

\newcommand{\mm}{\bar{M}}

\newcommand{\const}{\kappa}

\def\S{S}
\def\A{A}

\def\R{\mathbb{R}}
\def\E{\mathbb{E}}
\def\g{\gamma}
\def\l{\left}
\def\r{\right}

\begin{document}
\title{Bayesian regularization of empirical MDPs}

\author{\name Samarth Gupta\thanks{Work done while Samarth was interning at Amazon.com, Inc.} \email samarthg@andrew.cmu.edu \\
 \addr Carnegie Mellon University\\
 \AND
 \name Daniel N. Hill  \\
 \addr Amazon.com, Inc.\\
 \AND
 \name Lexing Ying \\
 \addr Stanford Unviersity\\
 \AND Inderjit Dhillon\thanks{Work done while Inderjit was affiliated with Amazon.com, Inc.} \\
 \addr University of Texas, Austin and Google  \\
}

\editor{No editors}
\maketitle

\begin{abstract}
  In most applications of model-based Markov decision processes, the parameters for the unknown
  underlying model are often estimated from the empirical data. Due to noise, the policy learned
  from the estimated model is often far from the optimal policy of the underlying model. When
  applied to the environment of the underlying model, the learned policy results in suboptimal
  performance, thus calling for solutions with better generalization performance.  In this work we
  take a Bayesian perspective and regularize the objective function of the Markov decision process
  with prior information in order to obtain more robust policies. Two approaches are proposed, one
  based on $L^1$ regularization and the other on relative entropic regularization. We evaluate our
  proposed algorithms on synthetic simulations and on real-world search logs of a large scale online
  shopping store. Our results demonstrate the robustness of regularized MDP policies against the
  noise present in the models.
\end{abstract}

\section{Introduction}

A Markov decision process (MDP) is a model $M = (S, A, P, r, \gamma)$. Here $S$ is the discrete
state space, with each state represented by $s$. $A$ represents the discrete action space with each
action denoted by $a$. $P$ denotes the transition probability tensor where, for each action $a \in
A$, $P^a \in\mathbb{R}^{|S| \times |S|}$ is the transition matrix between the states, i.e.,
$P_{st}^a$ denotes the probability of moving from state $s$ to state $t$ if action $a$ is taken at
state $s$. $r$ represents the reward tensor where, for each action $a \in A$,
$r^a\in\mathbb{R}^{|S|\times|S|}$ is the reward matrix between the states, i.e., $r_{st}^a$ denotes
the reward obtained in moving from state $s$ to state $t$ if action $a$ is taken at state $s$. The
discount factor $\gamma \in [0,1]$ determines the importance given to rewards obtained in the future
relative to those collected immediately.

A policy $\pi = (\pi_s^a)$ defines the probabilities of taking action $a$ at state $s$. The goal in
the MDP is to find a policy $\pi$ that maximizes the expected discounted cumulative reward $v_s^\pi$
for each state $s$, given by $v^\pi_s\equiv\E_\pi\l[\sum_{m=0}^\infty \g^m
  r^{a_m}_{s_m,s_{m+1}}|s_0=s\r]$. In what follows, $S$, $A$, and $\gamma$ are considered to be
fixed, and therefore, we often denote the MDP model in short with $M = (P,r)$.


For most applications, the environment is modeled with an unknown underlying MDP
$\bar{M}=(\bar{P},\bar{r})$ that is not directly accessible. The {\em empirical model} $M=(P,r)$ is
often estimated from samples of $\bar{M}=(\bar{P},\bar{r})$ and an optimal policy $\pi$ is then
learned from $M$. As the two models $M$ and $\bar{M}$ are different due to sampling noise, the
policy $\pi$ learned from $M$ is different from the true optimal policy $\bar{\pi}$ of $\bar{M}$.
When applying $\pi$ directly to the environment modeled by the underlying MDP $\bar{M}$, one often
experiences suboptimal performance.

To give a simple example, consider an MDP with only two actions $a_1$ and $a_2$ at each state and
that action $a_1$ is always better than $a_2$ for all states $s \in S$ under the true transition
probability matrices $\bar{P}^a$ and reward matrices $\bar{r}^a$. The transition matrices $P^a$ and
reward matrices $r^a$ constructed from samples are different from $\bar{P}^a$ and $\bar{r}^a$ due to
the noise present in the data. As a result, the policy $\pi$ learned from the empirical model
$M=(P,r)$ may recommend action $a_2$ over $a_1$ for some states $s \in S$, thus leading to poor
generalization performance.

As a more concrete example, let us consider the shopping experience of a customer at an online
shopping store. A customer starts a session by typing in an initial query. Based on the given query,
the store can recommend products using one of the existing search algorithms. Upon viewing the
results, the customer may either make a purchase or continue the browsing session by typing a
modified or new query. Such a shopping experience can be modeled as an MDP, where each query is
regarded as a state $s\in S$ and each search algorithm as an action $a\in A$. The reward $r_{s,t}^a$
corresponds for example to whether a purchase is made or not. The optimization problem is to decide
which search algorithm should be activated given the query in order to improve the overall shopping
experience. In a typical offline learning setting, the empirical MDP model $M$ is constructed based
on the historical log data. Therefore, the transition tensor $P$ and purchase actions are inherently
noisy. If one learns the policy directly from the noisy empirical model, it can have a poor
performance when deployed in the future.  While we motivate this issue using the online shopping
example, it exists universally in many other applications.



In this work, we study the problem of learning a robust MDP policy from the empirical model $M$ that
can perform significantly better than the naive policy $\pi$ from $M$ when deployed to the unknown
underlying $\bar{M}$. Though this is a challenging problem stated as it is, a key observation is
that in many real applications there is often prior information on the rankings of the available
actions. Here, we take a Bayesian approach to incorporate such prior information as a regularizer.


\paragraph{Main Contributions.} The main contributions of this work are:
\begin{itemize}
\item We propose a Bayesian approach that factors in known prior information about the actions and
  learns policies robust to noise in empirical MDPs.  More specifically, two approaches are proposed: one based on
  $L^1$ regularization and the other on relative entropic regularization. Both can be implemented
  efficiently by leveraging existing algorithms for MDP optimization.
\item We evaluate the designed algorithms on both synthetic simulations and on the logs of a
  real-world online shopping store dataset. Our regularized policies significantly outperform the
  un-regularized MDP policies.
\end{itemize}


\paragraph{Related work.}
When solving the MDPs, entropy regularization has proven quite useful
\cite{peters2010relative,fox2015taming,schulman2015trust,mnih2016asynchronous}. Commonly, Shannon
entropy or negative conditional entropy is used to regularize the MDPs
\cite{peters2010relative,fox2015taming,schulman2015trust,mnih2016asynchronous,dai2018sbeed,haarnoja2018soft}. While
this results in more robust stochastic policies, they do not necessarily account for any prior
information. The work of \cite{neu2017unified, peters2010relative, nachum2017trust, wu2019behavior} discusses relative entropic regularization in MDPs,
which biases results to a reference distribution. These works focus on improving the convergence and stability of RL methods by employing entropic regularization. But, this idea has yet to be applied in the
context of empirical MDPs through a Bayesian perspective. 

There has been work on reward shaping \cite{ng1999policy, harutyunyan2015shaping,
  cooper2012performance, grzes2017reward, gimelfarb2018reinforcement} where the idea is to obtain a
new MDP model $M’ = (P, r’)$ by modifying the rewards of model $M = (P, r)$ as $r'^{a}_{s,t}=
r_{s,t}^a + \phi(s) - \phi(t)$, where $\phi(s), \phi(t)$ are potential functions at state $s,t$. In
particular, \cite{ng1999policy} showed that such a reward shaping ensures that the optimal policy in
the two models $M$ and $M’$ remains the same. The focus of all these works is to design potential
functions $\phi$ to improve the convergence of algorithms in $M'$ without altering the optimal
policies. As the empirical model $M$ and true model $\bar{M}$ are different in our setting due to
the inherent noise, we need regularization based approaches which incorporate the prior information
about preference towards certain actions as the optimal policy in $M$ is not necessarily optimal
under $\bar{M}$.

An alternative solution to this problem would be from a denoising perspective, where the empirical
model $M$ is first denoised and then the policy is learned from the denoised model instead. This has
been studied in the context of linear systems where the objective is to solve the system
$\bar{A}x=\bar{b}$ but the estimated model parameters $A\approx\bar{A}$ and $b\approx\bar{b}$
contain a significant level of noise. To tackle this, \cite{etter2020operator,etter2021operator}
propose an operator augmentation approach that perturbs the inverse of the sampled operator
${A}^{-1}$ for better approximation to $x$. However, it is not clear how to extend this approach to
the control setting in MDP.

A closely related line of work is that of model based Bayesian reinforcement learning \cite{ghavamzadeh2016bayesian}, where priors are expressed over model as opposed to the policy. Imposing priors in such a way allows one to deal with imprecise models \cite{levine2020offline}. Our work on studying Bayesian regularization policies in the action space is complementary to this line of work. The choice of imposing a prior on model against a prior on policy boils down to the application domain. In several application domains, the state space is quite large as a result of which working with a Bayesian model on transition/reward tensors becomes infeasible. In contrast, the action space is relatively much smaller, as a result of which employing a Bayesian approach on the action space is much more practical.




\paragraph{Organization.}
In Section 2, we present the optimization formulations and describe our regularization approach by
incorporating prior information. Section 3 studies the performance of the proposed policy against
baseline algorithms on several simulated examples. Section 4 evaluates the performance of our
proposed algorithms on an application data set.

\section{Problem statement and algorithms}

\subsection{Policy maximization}

Let $ \Delta = \{\eta=(\eta^a)_{a\in\A}: \sum_{a\in\A} \eta^a = 1\text{ and } \eta^a \geq 0\}$ be
the probability simplex over the action set $A$. The set of all valid policies is
\[
\Delta^{|\S|} = \left\{\pi=(\pi_s)_{s\in\S}: \pi_s \in \Delta \text{ for }\forall s\in\S \right\}.
\]
For a policy $\pi \in \Delta^{|\S|}$, the transition matrix $P^\pi\in\R^{|\S|\times|\S|}$ under the
policy $\pi$ is defined as $P^\pi_{st} = \sum_{a\in\A}P^a_{st}\pi^a_s$, i.e., $P^\pi_{st}$ is the
probability of arriving at state $t$ from state $s$ if policy $\pi$ is taken. Similarly, the reward
$r^\pi\in\R^{|\S|}$ under the policy $\pi$ is given by $r^\pi_{s} = \sum_{a\in\A} r_s^a \pi^a_s$,
where $r_s^a = \sum_{t \in S} r^a_{st} P^a_{st}$, i.e., the expected reward at state $s$ under
action $a$.



For a discounted MDP \cite{sutton2018reinforcement, puterman2014markov} with $\g \in [0,1]$, the
value function under policy $\pi$ is a vector $v^\pi \in\R^{|\S|}$, where each entry $v^\pi_s$
represents the expected discounted cumulative reward starting from state $s$ under the policy $\pi$,
i.e.,
\[
v^\pi_s  = \E \l[\sum_{m=0}^\infty \g^m r^{a_m}_{s_m, s_{m+1}} | s_0 = s\r],
\]
with the expectation taken over $a_m\sim\pi_{s_m}$ and $s_{m+1}\sim P^{a_m}_{s_m,\cdot}$ for all
$m\geq 0$. The value function satisfies the Bellman equation \cite{bellman1966dynamic}, i.e., for
any $s\in\S$
\begin{equation*}
  v^\pi_s = r^\pi_s + \g \E^\pi [v^\pi_{s_1}|s_0 = s] = r^\pi_s + \g\sum_{t\in\S}P^\pi_{st}v^\pi_t,
\end{equation*}
or equivalently in the matrix-vector notation 
\begin{equation*}
  v^\pi = r^\pi + \g P^\pi v^\pi \quad \Longleftrightarrow \quad v^\pi = (I - \g P^\pi)^{-1} r^\pi.
\end{equation*}
Here, the inverse of the matrix $ (I - \g P^\pi)$ exists whenever $\g < 1$ or there exists a terminal
state $z$ in the MDP such that $P_{z,z}^a = 1$, $r_{z,z}^{a} = 0$, and
$P_{s,z}^a\neq 0\;\forall{a\in A,s\in S}$ \cite{bell1965gershgorin}.  Given the MDP, the optimization problem is
\begin{equation}\label{eqn:gn_p}
  \max_{\pi} \  e^\intercal v^\pi = \max_{\pi}  e^\intercal (I - \g P^\pi)^{-1} r^\pi,
\end{equation}
where $e\in\R^{|\S|}$ is an arbitrary vector with positive entries \cite{ye2011simplex}. By
introducing the discounted visitation count $w^\pi = (I - \g P^\pi)^{-\intercal} e$, we can rewrite
\Cref{eqn:gn_p} as
\begin{equation}\label{eqn:gn_q}
  \max_{\pi} \  (w^\pi)^\intercal r^\pi,
  \quad\text{with}\quad
  w^{\pi} \equiv  (I - \g P^\pi)^{- \intercal} e.
\end{equation}


\subsection{Bayesian approaches}

Given an MDP model $M$, we can view \eqref{eqn:gn_q} as the maximum a posteriori probability (MAP)
estimate $\max_\pi \Pr(\pi | M)$ with
\begin{equation}
  \Pr(\pi | M)
  \propto \exp(e^\intercal v^\pi)
  = \exp((w^\pi)^\intercal r^\pi).
\label{eqn:gen}
\end{equation}
When prior knowledge about $\pi$ is not available, it is natural to take a uniform prior over $\pi$,
i.e. $\Pr(\pi)$ is constant. This implies that $\Pr(\pi|M)\propto\Pr(M|\pi)\Pr(\pi)\propto\Pr(M|\pi)$,
leading to
\begin{equation}
  \Pr(M | \pi) \propto \exp((w^\pi)^\intercal r^\pi).
  \label{eqn:Mpi}
\end{equation}


On the other hand, if prior knowledge about $\pi$ is available, it makes sense to impose more
informative priors on $\pi$. One commonly used prior is that at state $s\in S$ an action $\xi(s)\in
A$ is often preferred over the rest of the actions. This prior information can be incorporated
naturally in the following two ways.


\subsubsection{$L^1$-type prior}
\label{subsec:l1}
In particular, we assume a prior 
\begin{equation}
  \Pr(\pi) \propto  \exp(- \lambda (w^\pi)^\intercal f^\pi),
  \label{eqn:prior}
\end{equation}
where $f^\pi$ is defined to be the $L^1$ norm of $\pi$ outside of action $\xi(s)$, i.e.,
$(f^\pi)_s=\sum_{a \neq \xi(s)} \pi_{s,a}$. This prior puts more probability on action $\xi(s)$
relative to other actions in $A$. Combining \eqref{eqn:Mpi} and \eqref{eqn:prior} leads to the a
posteriori probability
\begin{align*}
  \Pr(\pi | M) &\propto \Pr(\pi, M) = \Pr(M| \pi) \Pr(\pi) \\
  &\propto \exp( (w^\pi)^\intercal r^\pi) \exp(-  \lambda (w^\pi)^\intercal f^\pi).
\end{align*}
The corresponding MAP estimate is
\begin{align}
  &   \arg\max_\pi \;\exp((w^\pi)^\intercal r^\pi - \lambda (w^\pi)^\intercal f^\pi) \nonumber \\
  = & \arg\max_\pi \;(w^\pi)^\intercal (r^\pi - \lambda f^\pi) 
  \label{eqn:l1argmax}
\end{align}
Note that individual component of $r^\pi - \lambda f^\pi$ can be broken down as, 
\[
r^\pi_s - \lambda f^\pi_s = \sum_a r_s^a \pi_s^a - \lambda \sum_{a \neq \xi(s)} \pi_s^a = \sum_a (r_s^a -
\lambda \delta_{a \neq \xi(s)}) \pi_s^a,
\]
where $\delta_{a \neq \xi(s)}=1$ if $a\neq\xi(s)$ and $0$ if $a=\xi(s)$. Hence, this formulation is equivalent to replacing $r_s^a$ with
$r_s^a-\lambda\delta_{a\neq\xi(s)}$, i.e., the reward of all non-preferred actions $a \neq\xi(s)$ is
reduced by a constant $\lambda$. The corresponding Bellman equation is
\[
v_s = \max_a ((r_s^a - \lambda \delta_{a \neq\xi(s)}) + \g (P^a v)_s ).
\]
Once $v_s$ is computed, the optimal action at state $s$ is given by
\begin{equation}
  \arg \max_a ((r_s^a - \lambda \delta_{a \neq\xi(s)}) + \g (P^av)_s).
  \label{eqn:l1res}
\end{equation}

\subsubsection{Relative entropy regularization}
\label{subsec:relEnt}
In the MDP literature, it is common to use Shannon entropy regularization, which allows for learning
stochastic policies instead of deterministic ones. However, it fails to capture the prior
information, such as the scenario where one of the actions in $A$ is preferred over others. To
accommodate such a prior, we propose to use the relative entropy instead of Shannon entropy. By
choosing the prior distribution carefully, relative to which the entropy of policy is evaluated, we
obtain solutions that prefer one action over other actions in $A$. We consider a penalty such that
$\Pr(\pi) \propto \exp(- \const (w^\pi)^\intercal h^\pi)$ with
$h^\pi_s=\sum_{a}\pi^a_s\log\left(\frac{\pi^a_s}{q^a_s}\right)$, where $q^a_s$ is a distribution
over $A$ that prefers $a=\xi(s)$, e.g.,
\[
q_a=
\begin{cases}
  1 - \epsilon,       & a=\xi(s)\\
  \epsilon/(|A| - 1), & a\neq\xi(s),
\end{cases}
\]
for some $\epsilon > 0$. The corresponding MAP estimate is 
\begin{align}
  &\arg\max_\pi \;(w^\pi)^\intercal r^\pi - \const (w^\pi)^\intercal h^\pi \nonumber \\
  = &\arg\max_\pi \sum_s w_s^\pi\left(r_s^\pi - \const \sum_a \pi_s^a \log \frac{\pi_s^a}{q_s^a} \right) \nonumber \\
  = &\arg\max_\pi \sum_s w_s^\pi\left(\sum_a ( (r_s^a + \const \log q_s^a) - \const  \log \pi_s^a ) \pi_s^a  \right) \label{eqn:regMDP}
\end{align}
This is in fact equivalent to the standard Shannon entropy regularization with modified rewards
$r_s^a+\const \log q_s^a$, i.e., a penalty $\const \log q_s^a$ is added to $r_s^a$ when action $a
\in A$ is taken. The magnitude of the penalty for action $a$ is large if the prior probability
$q_s^a$ of selecting action $a$ is small. By applying the Bellman equation of Shannon entropy
regularization \cite{neu2017unified,ying2020note} to \eqref{eqn:regMDP}, we obtain
\begin{align*}
  v_s = \max_{\pi_s\in \Delta} \sum_{a\in\A}(r^a_s + \const \log q_s^a + & \g \sum_{t\in\S}P^a_{st} v_t - \kappa \log \pi_s^a)\pi^a_s.
\end{align*}
Because of the Gibbs variational principle, the RHS is equal to \\ $\const \log\left(\sum_{a\in\A} \exp
\left(\frac{r_s^a + \const \log q_s^a + \g \sum_{t\in\S}P^a_{st}v_t}{\const}\right)\right)$. Thus,
the Bellman equation can be written in the following log-sum-exp form
\begin{equation}\label{eqn:gr_logsum}   
  v_s = \const \log\left(\sum_{a\in\A} \exp \left(\frac{r_s^a + \const \log q_s^a + \g \sum_{t\in\S}P^a_{st}v_t}{\const}\right)\right),
\end{equation}
which can be solved with a value function iteration. Once $v_s$ is known, the optimal policy at
state $s$ and action $a$ is given by
\begin{equation}  
  \frac{\exp(r^a_s+ \const \log q_s^a + \g\sum_{t\in\S} P^a_{st}v_t-v_s)}{Z_s},
  \label{eqn:erres}
\end{equation}
where $Z_s = \sum_{a\in\A} \exp(r^a_s+ \const \log q_s^a + \g\sum_{t\in\S} P^a_{st}v_t-v_s)$ is the
normalization factor.

\subsubsection{Comments}
Both the optimization formulations in Sections \ref{subsec:l1} and \ref{subsec:relEnt} add a penalty
on top of the reward obtained for each action. The less preferred actions (i.e., $a \neq \xi(s)$)
are penalized and hence as a result the learned policy prefers action $\xi(s)$ over the other
ones. The magnitude of the penalty depends on the regularization parameter $\lambda$ in the $L^1$
case and $(\const,q^a)$ in the relative entropy case. The policy obtained in \Cref{subsec:l1} is a
deterministic policy, whereas the one learned in \Cref{subsec:relEnt} is a stochastic policy due to
the added entropy regularization. When $\const$ in \Cref{subsec:relEnt} is chosen to be small, the
policy becomes more and more concentrated and is often practically equivalent to the one in
\Cref{subsec:l1}. Finally, we note that both the approaches presented above can be easily extended
to settings where a certain subset of the actions are preferred over the others.


\section{Simulated examples}

In the following simulated examples, we demonstrate numerically that an optimal policy of the
empirical MDP $M$ results in sub-optimal performance on the underlying MDP model $\bar{M}$ and that
the regularized policies provide significantly better performance.

\begin{figure}[h]
  \centering
  \includegraphics[width = 0.48\textwidth]{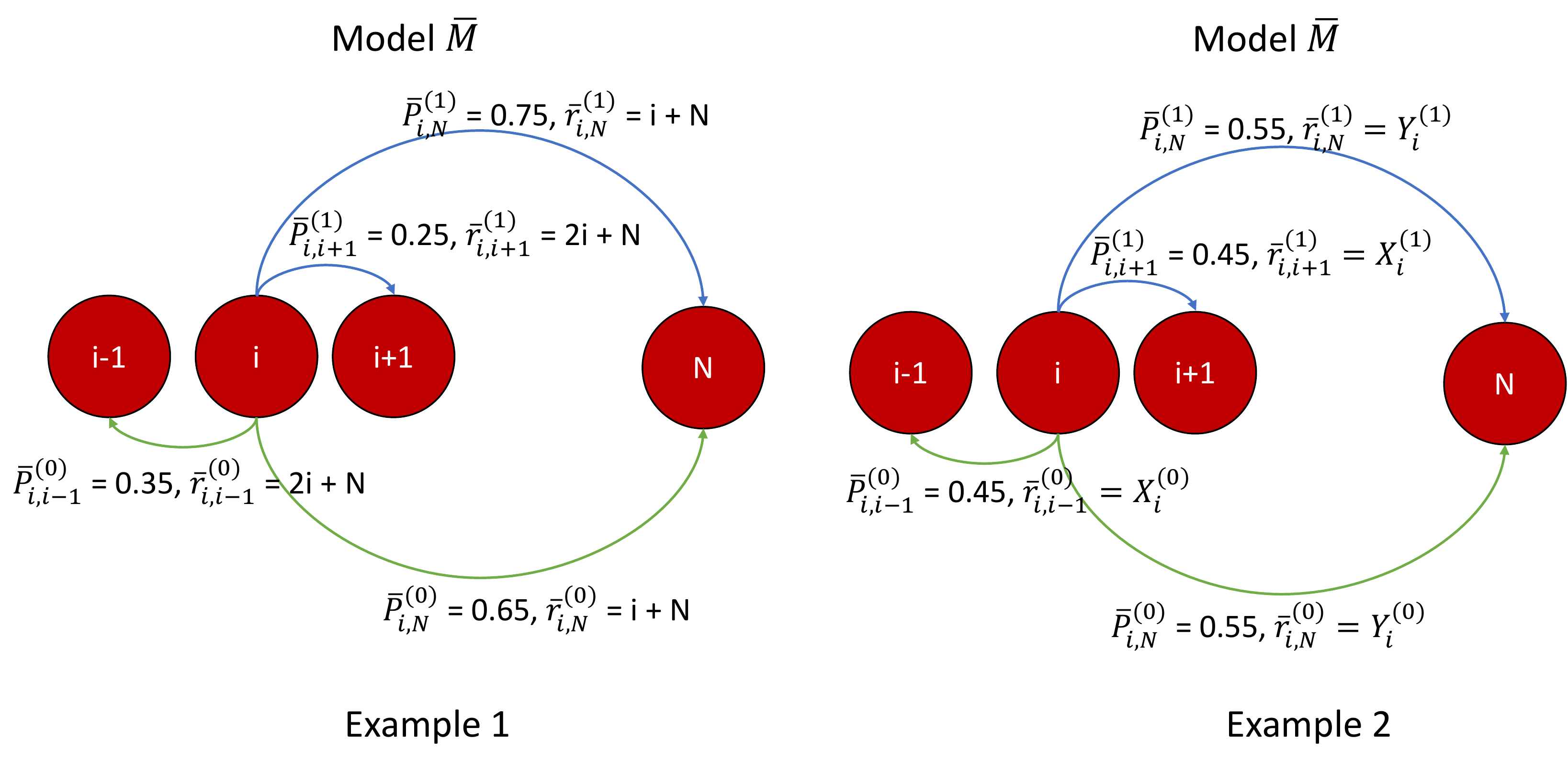}
  \caption{
    The transition probabilities $\bar{P}_{st}^a$ and rewards $\bar{r}_{st}^a$ of the true
    model $\bar{M}$ in Example 1 (left) and Example 2 (right). The state space is the set
    $\{1,2,3,\ldots,N\}$. Under action $0$, for $i \neq 1$ the transition from state $i$ can go to
    state $i-1$ or $N$, while for $i = 1$ it can go to $N-1$ or $N$. Similarly under action $1$, for
    $i \neq N - 1$ the transition from state $i$ can go to $i + 1$ or $N$, while for $i = N-1$ it
    can go to state $1$ or $N$.  }
  \label{fig:trainModel1}
\vspace{2mm}
\end{figure}


\paragraph{Example 1.}
Consider the MDP model $\bar{M}$ shown in \Cref{fig:trainModel1}. This model has $N$ states
$\{1,2,\ldots,N\}$ and two actions $\{0,1\}$. The transition and reward tensors for model $\bar{M}$
are defined below
\[
\left\{
\begin{aligned}
\bar{P}^{(0)}_{i, N - 1 - (N - i)\%(N-1) } &= 0.35,  &\bar{P}^{(0)}_{i,N} &= 0.65 \\
\bar{r}^{(0)}_{i, N - 1 - (N - i)\%(N-1)  } &= 2i + N, &\bar{r}^{(0)}_{i,N} &= i + N \\
\bar{P}^{(1)}_{i, i \% (N - 1) + 1} &= 0.25,  &\bar{P}^{(1)}_{i,N} &= 0.75 \\
\bar{r}^{(1)}_{i, i \% (N - 1)  + 1} &= 2i + N,    &\bar{r}^{(1)}_{i,N} &= i + N. \\
\end{aligned}
\right.
\] 
Here, state $N$ is a terminal state, where $\bar{P}^{0/1}_{N,N} = 1$ and $\bar{r}^{0/1}_{N,N}=0$. In
our simulations, $N = 10$ and $\gamma=1$. The empirical MDP $M$ is constructed from the true MDP
$\bar{M}$ by sampling the transition probabilities from a set of 100 samples for each state $s \in S$. 
Due to the sampling noise, the transition tensor $P$ of model $M$ is different from
$\bar{P}$ of $\bar{M}$. In this example, we assume that the underlying reward tensor $\bar{r}$ is
known exactly. Under the true model $\bar{M}$, action $\xi(s)\equiv 0$ is optimal for any state
$s\in S$, as the transition probability to states with higher reward is larger under action
$0$. However, when the model $M$ is constructed from empirical samples, the optimal policy learned
on $M$ recommends action $1$ for some states, leading to a sub-optimal performance on the true model
$\bar{M}$.


\begin{figure}[h]
    \centering
    \includegraphics[width = 0.48\textwidth]{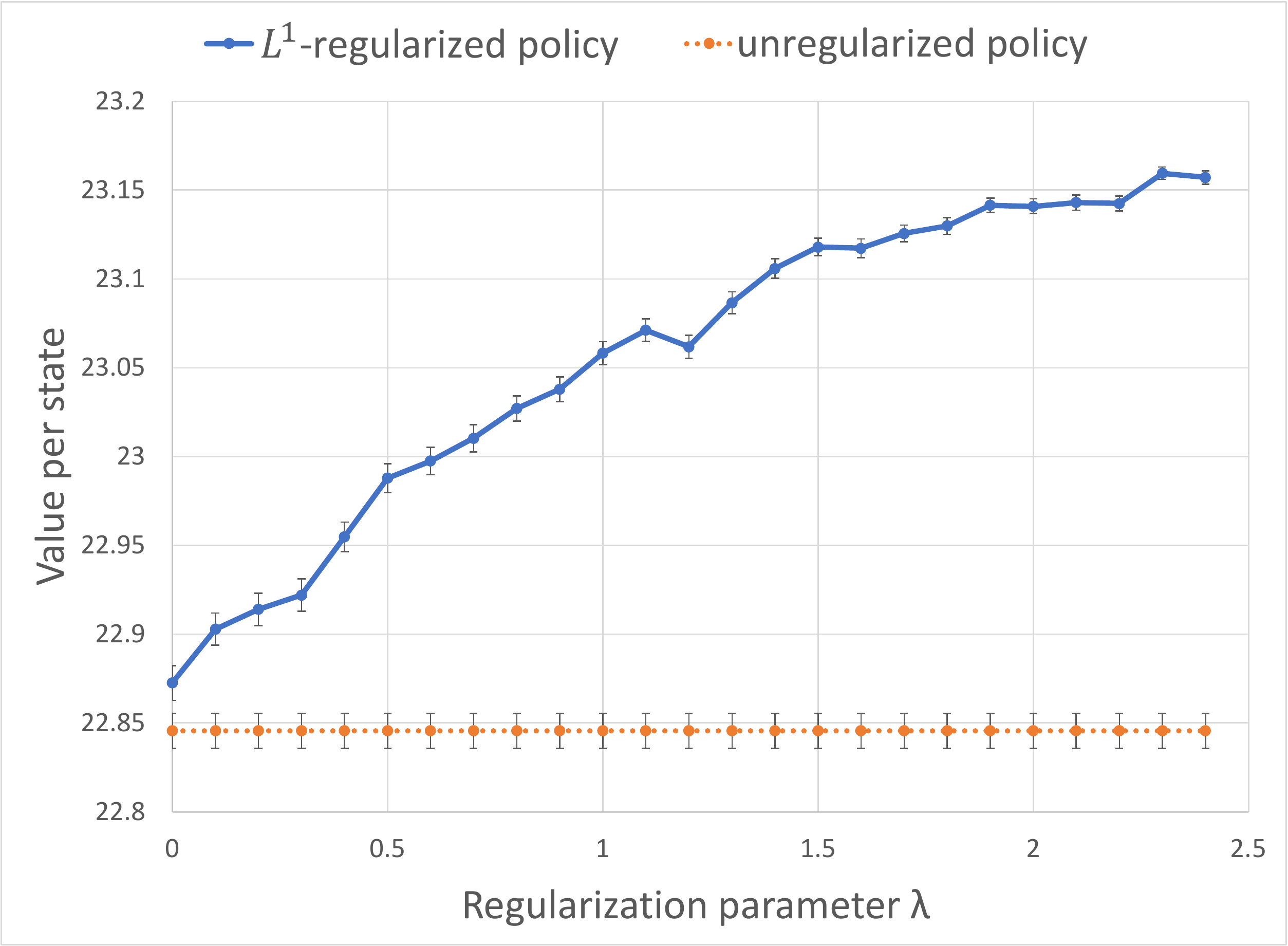}
    \caption{
      Example 1. The $L^1$-regularized policy vs. the unregularized policy. When $\lambda$
      increases, the $L^1$-regularized policy prefers action $0$ over action $1$ and achieves
      significantly better value per state on the true model $\bar{M}$ relative to the unregularized
      policy of model $M$. The error bars indicate a width of two standard error in all subsequent
      simulations and experiments.}
    \label{fig:sim1_reg}
\vspace{2mm}
\end{figure}


\begin{figure}[h]
    \centering
    \includegraphics[width = 0.48\textwidth]{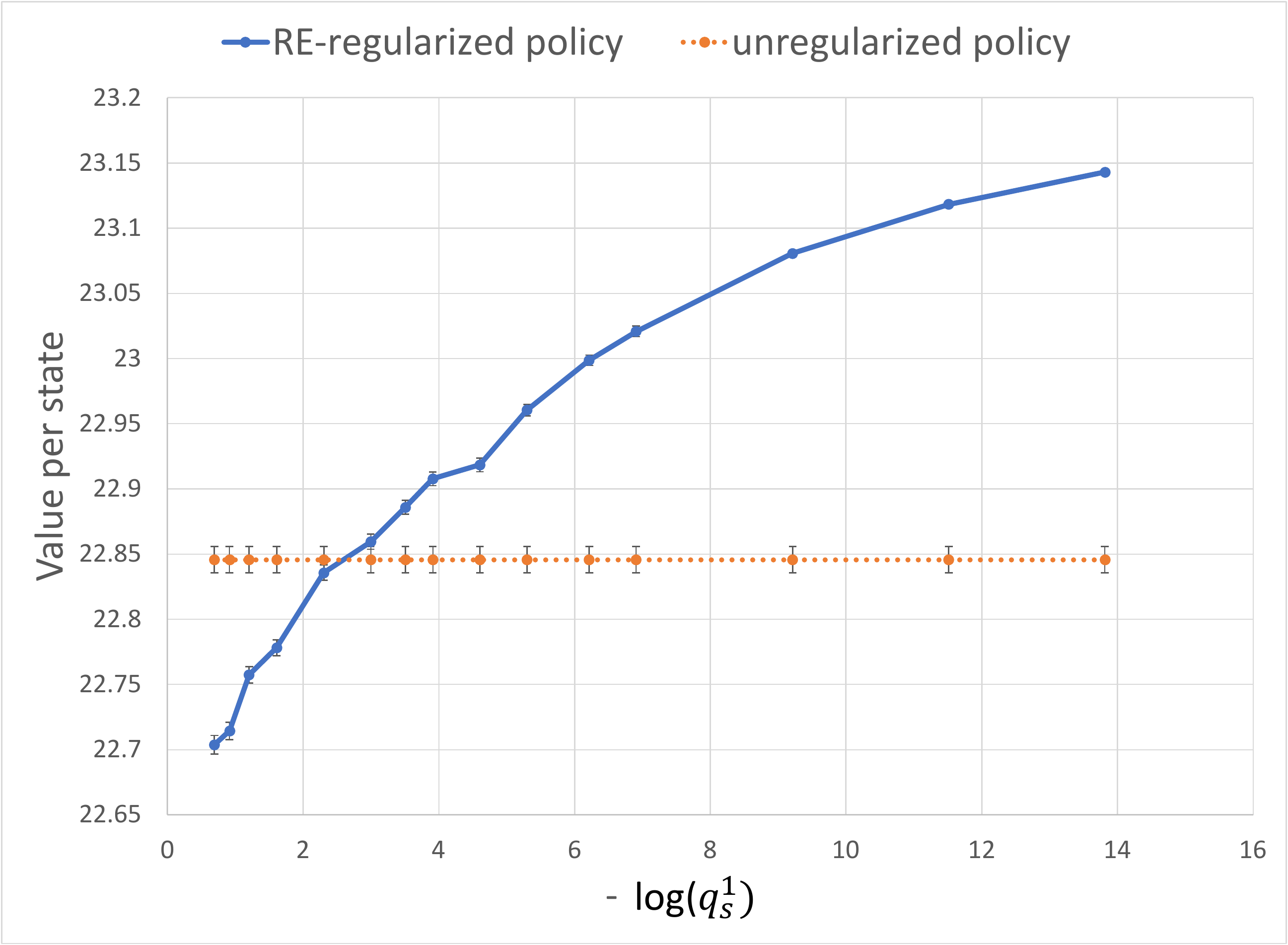}
    \caption{ Example 1. The RE-regularized policy vs. the unregularized policy.  As the value of
      $q_s^{1}$ decreases (or equivalently $-\log(q_s^1)$ increases), the RE-regularized policy
      prefers action $0$ over action $1$, leading to improvement in performance over the
      unregularized policy.  }
    \label{fig:sim1_rel}
\vspace{2mm}
\end{figure}

\Cref{fig:sim1_reg} shows the comparison between the $L^1$-regularized policy and the unregularized
policy when evaluated on $\bar{M}$. As the regularization parameter $\lambda$ increases, the learned
policy prefers action $0$ over action $1$ and obtains higher value per state as a result.  In
\Cref{fig:sim1_rel}, we compare the relative entropic regularization policy (referred in short as
the {\em RE-regularized policy}) with the unregularized policy on the true model $\bar{M}$. We set
regularization coefficient to be $\const = 0.25$ and vary the value of prior $q_s^1$ on the
$x$-axis.  The value $q_s^1 = 0.5$ corresponds to the case with Shannon entropic regularization. As
the value of $q_s^1$ becomes smaller, the RE-regularized policy prefers action $0$ over action $1$
and results in a higher value per state relative to the unregularized policy.

\begin{figure}[h]
    \centering
    \includegraphics[width = 0.48\textwidth]{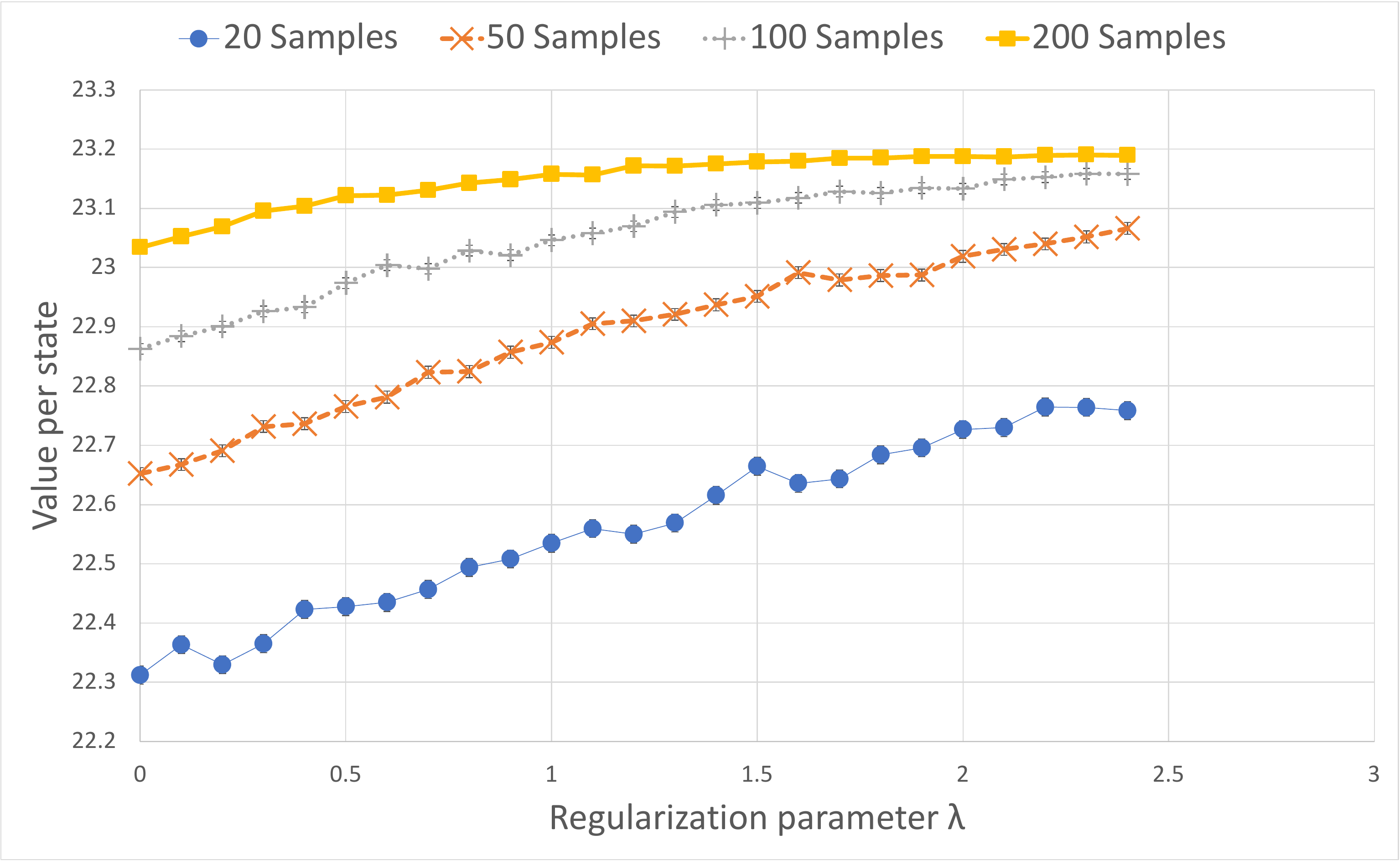}
    \caption{ Example 1. The performance of the $L^1$-regularized policy as a function of the
      samples used to estimate the transition probabilities for each state in $\bar{M}$. The regularization benefit decreases as the number of
      samples increases due to a reduction of the sampling noise.}
    \label{fig:sim1L1}
\vspace{2mm}
\end{figure}

In practice, the values of $(\const, q_s^{1})$ (for the RE-regularized policy) and $\lambda$ (for
the $L^1$-regularized policy) can be learned through evaluation on a validation set. In both the
cases, the need for regularization goes down as the number of samples used to evaluate $\bar{M}$
increases. This effect is demonstrated in \Cref{fig:sim1L1}, where we plot the performance of the
$L^1$-regularized policy as a function of the samples used to estimate the transition probabilities for each state in 
$\bar{M}$.

%


\paragraph{Example 2.}
In Example 1, only the transition tensor in $M$ is sampled from the true model $\bar{M}$. In
practice, the reward tensor is also estimated empirically. To model this, in this example the reward
values $r_{st}^a$ has a Gaussian noise $\mathcal{N}(0, \sigma^2)$ added to the true unknown
underlying reward $\bar{r}_{st}^a$. The transition and reward tensors for model $\bar{M}$
are defined below (also see \Cref{fig:trainModel1})
\[
\left\{
\begin{aligned}
\bar{P}^{(0)}_{i, N - 1 - (N - i)\%(N-1) } &= 0.45,  & \bar{P}^{(0)}_{i,N} &= 0.55 \\
\bar{r}^{(0)}_{i, N - 1 - (N - i)\%(N-1)  } &= X_i^{(0)}, &\bar{r}^{(0)}_{i,N} &= Y_i^{(0)} \\
\bar{P}^{(1)}_{i, i \% (N - 1) + 1} &= 0.45,  &\bar{P}^{(1)}_{i,N} &= 0.55 \\
\bar{r}^{(1)}_{i, i \% (N - 1)  + 1} &= X_i^{(1)},    &\bar{r}^{(1)}_{i,N} &= Y_i^{(1)}. \\
\end{aligned}
\right.
\] 


Here, $X_i^{(0)}$ is taken to be a random realization drawn from $\mathcal{N}(5,1)$. Similarly,
$X_i^{(1)} \sim \mathcal{N}(6,1)$, $Y_i^{(0)} \sim \mathcal{N}(2,1)$,
$Y_i^{(1)}\sim\mathcal{N}(3,1)$. In our simulations, $N = 1000$. For this MDP model $\bar{M}$, the
optimal action is $0$ for about $82.4\%$ of the states in $\mm$ as the expected number of steps to
reach the terminal state $N$ from a given state $s$ is higher in action $0$ relative to action $1$.

\begin{figure}[h]
    \centering
    \includegraphics[width = 0.48\textwidth]{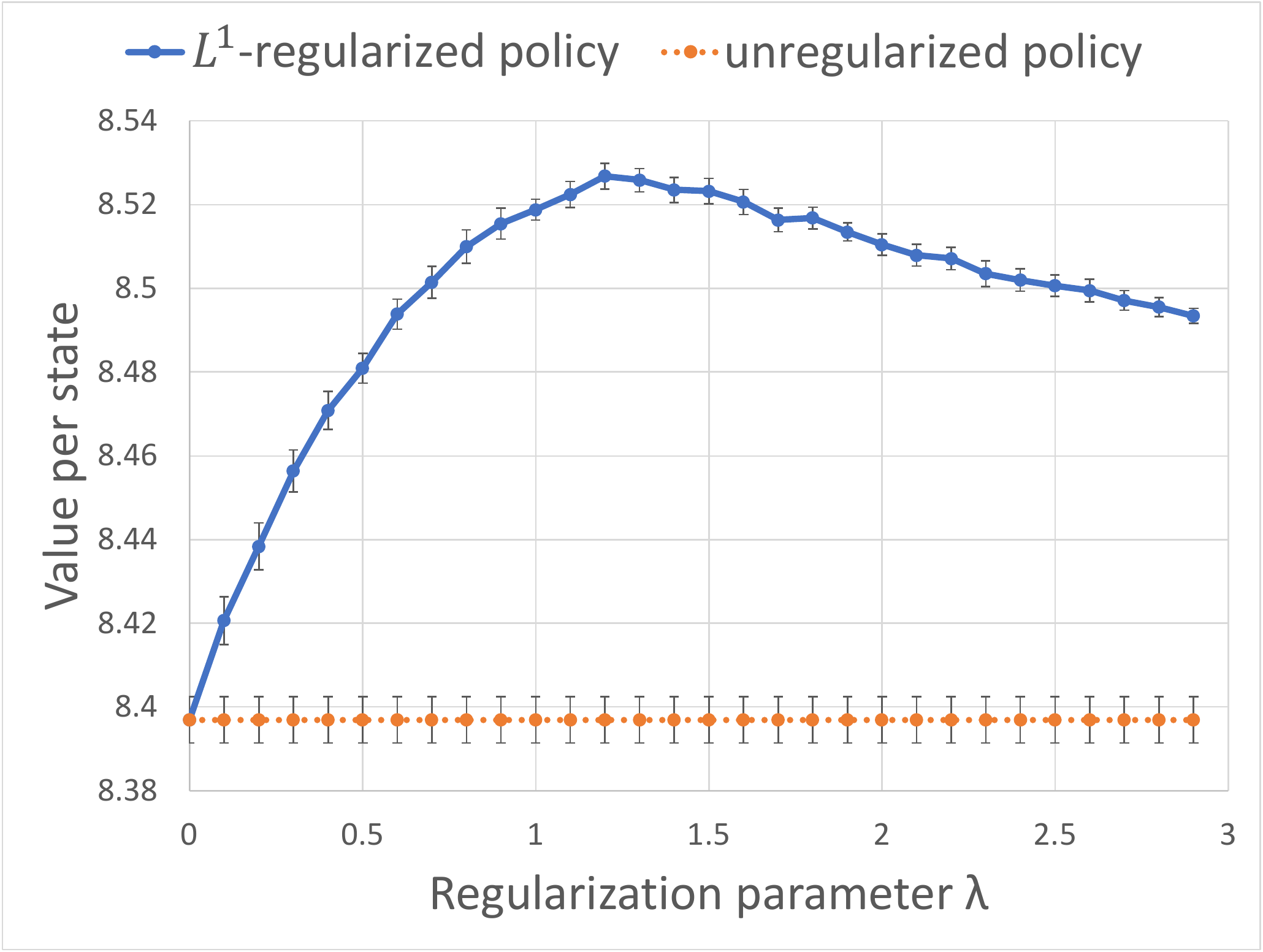}
    \caption{Example 2. The $L^1$-regularized policy vs. the unregularized policy.  As $\lambda$
      increases, the $L^1$-regularized policy favors action $0$ over action $1$. The incorporated
      prior information allows the $L^1$-regularized policy to outperform the unregularized
      policy. }
    \label{fig:sim2_reg}
\vspace{2mm}
\end{figure}

\begin{figure}[h]
    \centering
    \includegraphics[width = 0.48\textwidth]{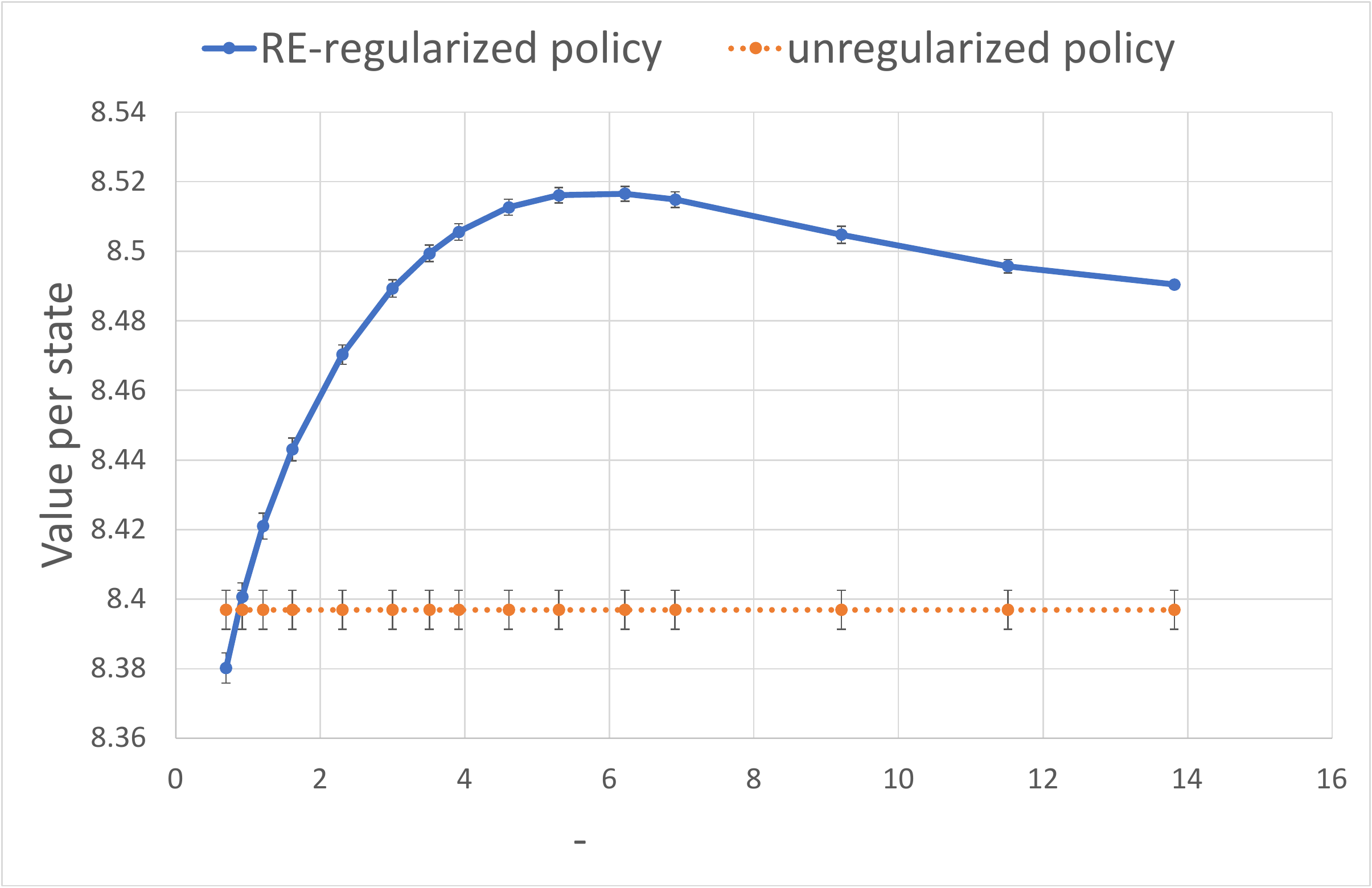}
    \caption{Example 2. The RE-regularized policy vs. the unregularized policy. As $q_{s}^1$
      decreases, the value of $-\log(q_s^1)$ increases and the RE-regularized policy favors action
      $0$ over action $1$. Accounting for prior information through $q_s^1$ helps the RE-regularized
      policy to outperform the unregularized policy. }
    \label{fig:sim2_rel}
\vspace{2mm}
\end{figure}

The empirical MDP $M$ is obtained by averaging $100$ samples per state of $\bar{M}$, where each
reward entry of $r$ is corrupted by a zero mean Gaussian noise with standard deviation
$\sigma=1.5$. As the transition and reward tensors in $M$ contain noise, the unregularized policy
from $M$ recommends action $1$ for more than $30\%$ of the states. As a result, the unregularized
policy is sub-optimal on the underlying model $\bar{M}$.

The prior information that action $0$ is preferred over action $1$ helps the $L^1$-regularized
policy and RE-regularized policy to outperform the unregularized policy. Figures \ref{fig:sim2_reg}
and \ref{fig:sim2_rel} illustrate this improvement as a function of regularization parameters
$\lambda$ and $q_s^{1}$ (the regularization coefficient $\const$ is fixed at $0.25$ for this
example). As the value of $\lambda$ or $-\log(q_s^1)$ increase, the regularized policies favor
action $0$ over action $1$, leading to significant improvements on model $\bar{M}$. With further
increase in $\lambda$ and $-\log(q_s^1)$, the performance dips afterwards as the regularized
policies start selecting action $0$ over action $1$ for more states than necessary.

\section{Experiments on real data}

This section discusses large-scale experiments on logs of an online shopping store with competing
search algorithms. We consider the user shopping experience discussed in Section 1 and model a
shopping session with an MDP where a state $s \in S$ corresponds to the search query typed in by the
user.


\begin{figure}[h]
    \centering
    \includegraphics[width = 0.48\textwidth]{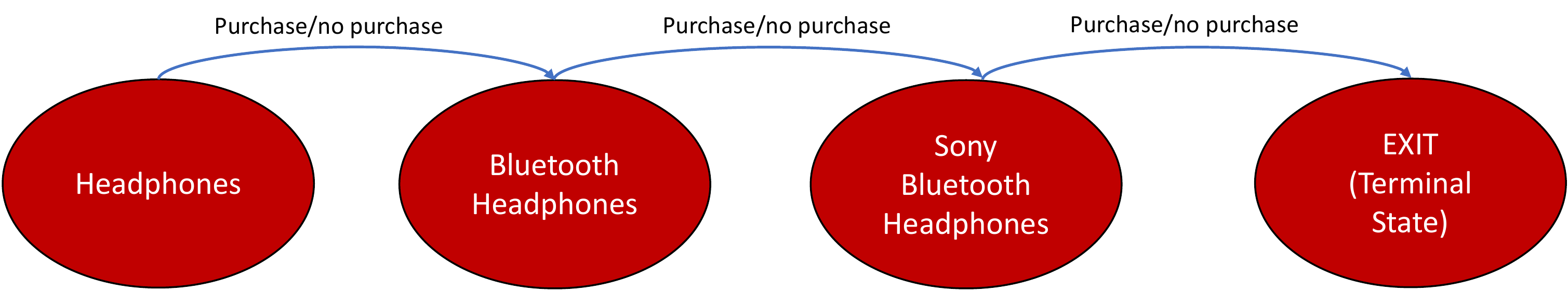}
    \caption{A shopping session at an online shopping store is typically sequential in nature. The
      user may start with a broad query and continuously refine it based on the results generated
      from the search algorithm. Eventually, the user exits the system, which is modeled by the
      terminal state. The user can make a purchase at any point during the session.}
    \label{fig:session}
\vspace{2mm}
\end{figure}

For each query $s$, the shopping store needs to decide on the search algorithm to use to display
results. This corresponds to the two actions of the MDP. When an action is taken, the search results
are shown to the user and the user interaction will result in a transition to a new state. We
identify this new state with the new query $t$ from the user. However, before making this transition
from state $s$ to $t$, the user may make a purchase, which corresponds to the reward. The user may
terminate the session at any point with/without making a purchase and this is captured by the
transition to a terminal state (see \Cref{fig:session}). The rewards are considered to be binary: if
a user makes a purchase at state $s$ under shopping store's action $a$ and then transitions to $t$, $r_{s,t}^{a} = 1$. $r_{s,t}^{a} = 0$ if no purchase were made. The two available search algorithms perform differently on
different queries. Therefore, there is an opportunity to interleave different algorithms based on
the queries, even within a single shopping session. Moreover, often it is known a priori that one
search algorithm may work better than the other. As a result, it is useful to incorporate this
information as a prior and design regularized policies that are robust to noise in the empirical
MDP.
%


To conduct our experiment, we collected the search logs of an online shopping store for a time period for two different search algorithms, one deployed in period 1 and the other in period 2, with period 1 is before period 2 in time and both the time periods are non-overlapping. Therefore, the action space consists of two search algorithms, $A =\{ranker 1, ranker 2\}$ which were previously deployed in period 1 and period 2 respectively. We processed the data in a
time period where ranker 1 was deployed to obtain the search logs under the ranker 1 action. Similarly, the data for another time period, in which ranker 2 was used, was collected to obtain the user logs under the ranker 2 action. We
considered the set of $135,000$ most typed queries as the state space $S$. For each of these time
periods, we estimated the transition and reward tensors from user logs, thereby obtaining the MDP
model $M$, i.e., $P_{s,t}^{algo1}$, $P_{s,t}^{algo2}$, $r_{s,t}^{algo1}$, $r_{s,t}^{algo2}$ for all $s,t$.

The key challenge is to learn robust policies from the empirical model $M$. It is known a priori
that on average, ranker 2 tends to produce better results relative to ranker 1. We exploit this information to learn the $L^1$-regularized and RE-regularized policies,
which interleave ranker 1 and ranker 2 effectively for different queries within a single
session.

The performance of different policies is judged based on the objective function $e^\intercal
v\equiv\sum_{s\in S} e_s v_s$, where $v_s$ is the value function at $s \in S$ with discount factor $\gamma = 1$ and $e_s$ denotes the
probability that $s$ is the first query in a random shopping session. This probability $\{e_s\}$ is
evaluated based on a hold-out time period from the collected search logs.


In order to evaluate the performance in an unbiased way, we extracted the data for different time
periods in which ranker 1 and ranker 2 were deployed, to construct a model $\tilde{M}$ with tensors denoted by $\tilde{P}$ and
$\tilde{r}$.  As the true underlying model $\bar{M}$ is not directly accessible, we use $\tilde{M}$,
a fresh unbiased estimator of $\bar{M}$, to evaluate the different policies. This essentially
corresponds to evaluating the performance of the policies on a new time period.

The transition and reward tensors between models $M=(P,r)$ and $\tilde{M}=(\tilde{P},\tilde{r})$ can
be quite different. In fact, comparing these two estimated models provides an idea of the existing
noise in the estimated models. For example, the average $L^1$ norm of the rows of
$P^{ranker1}-\tilde{P}^{ranker1}$ (also $P^{ranker2}-\tilde{P}^{ranker2}$) is about $0.16$, suggesting that
the average total variational distance between transition probability vectors of $P$ and $\tilde{P}$
is $0.08$, which is empirically quite significant.

For comparison purposes, we include the performance of several baselines defined below:
\begin{itemize}
\item unregularized MDP policy: this policy is optimal for model $M$ and is applied to
  $\tilde{M}$ without any regularization.
\item one-shot policy (OSP): this policy selects action ranker 1 for a particular keyword $s$ if
  immediate reward from ranker 1 is larger, i.e., $r_s^{ranker1} > r_s^{ranker2}.$
\item regularized one-shot policy ($\text{OSP}_\lambda$): this policy selects ranker 1 for a
  particular keyword $s$ only if $r_s^{ranker1} - \lambda > r_s^{ranker2}.$
\end{itemize}

\begin{figure}[t]
    \centering
    \includegraphics[width = 0.48\textwidth]{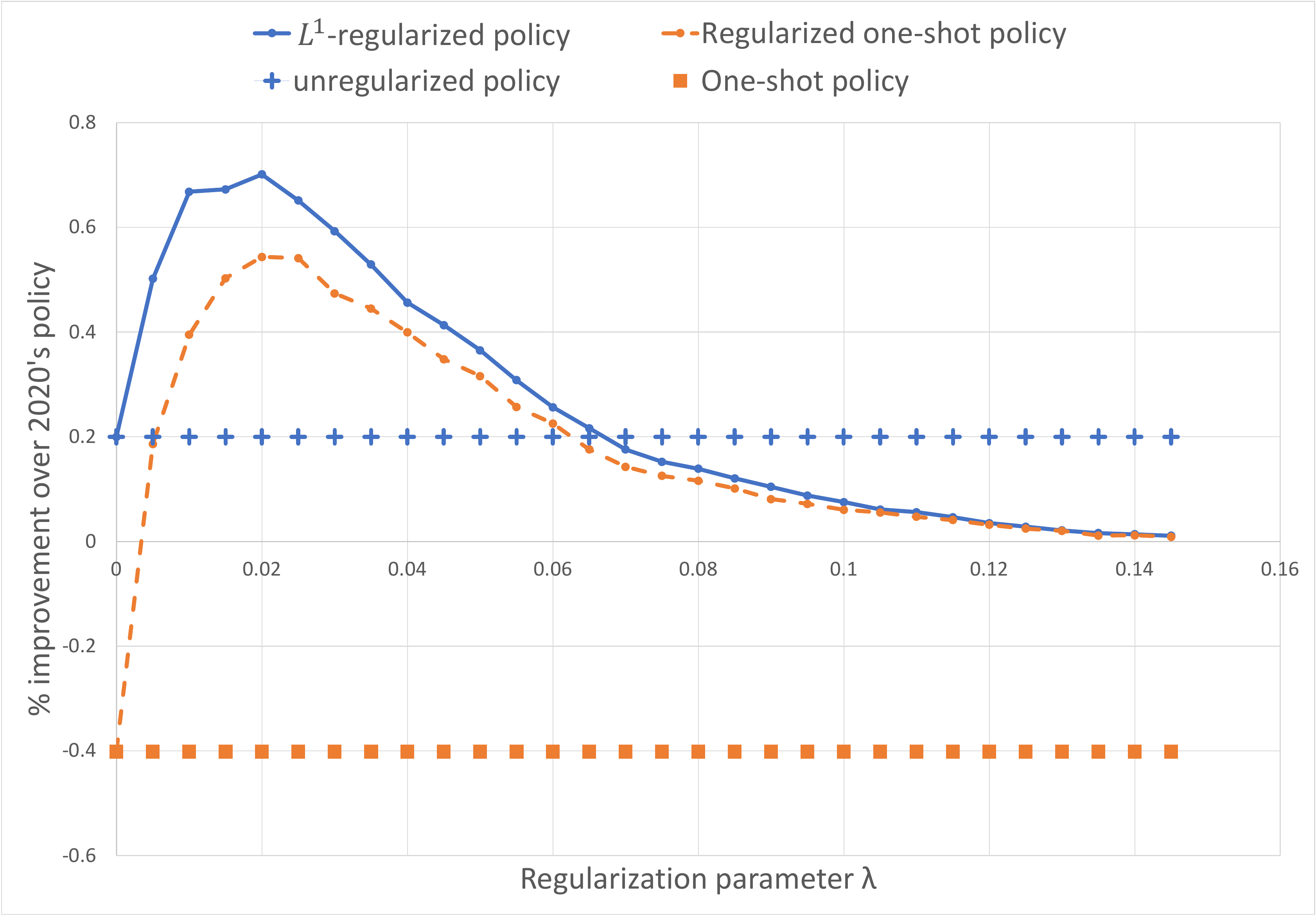}
    \caption{Performance comparison on an online shopping store dataset: The $L^1$-regularized
      policy accounts for the prior information through $\lambda$. As $\lambda$ increases, the
      $L^1$-regularized policy favors action $0$ over action $1$. The incorporated prior allows the
      $L^1$-regularized MDP policy to outperform the unregularized MDP policy learned from $M$.  Since the
      MDP based approach accounts for the delayed rewards by modeling the session interaction, it
      outperforms the regularized one-shot policy $\text{OSP}_\lambda$ for all values of $\lambda$.}
    \label{fig:exp1}
\vspace{2mm}
\end{figure}

\Cref{fig:exp1} shows the performance of $L^1$-regularized MDP policy, the unregularized MDP policy, OSP,
and $\text{OSP}_\lambda$ on model $\tilde{M}$, where all these policies are learned from model
$M$. We observe that (a) the $L^1$-regularized MDP policy shows about $0.7 \%$ improvement over the ranker 2 policy,
and the improvement over the ranker 1 policy is in the range of $21$-$22 \%$, (b) the $L^1$-regularized policy outperforms both OSP and
$\text{OSP}_\lambda$, suggesting that the MDP model is beneficial as the OSP and
$\text{OSP}_\lambda$ do not factor in delayed rewards, and (c) the $L^1$-regularized policy
outperforms the unregularized MDP policy by accounting for prior information in the form of
regularization parameter $\lambda$. The best performance is obtained with $\lambda = 0.02$.  As the
value of $\lambda$ increases, the regularized policies prefer ranker 2 over ranker 1. When
$\lambda$ is increased further, the learned policy ends up selecting ranker 2 for all the states.
This is why its performance becomes similar to ranker 2 in \Cref{fig:exp1} for large values
of $\lambda$.

The same experiment is repeated for the RE-regularized policy and we observe a performance similar to that of the $L^1-$regularized policy with the best
improvement of $0.69\%$ coming at $\const = 0.001$ and $q_s^{ranker1} = 10^{-8}$.


The above experiments suggest that hyperparameters $\lambda = 0.02$ and $(\const=0.001,
q_s^{ranker1}=10^{-8})$ perform the best for the $L^1$-regularized policy and the RE-regularized
policy, respectively. To validate our results, we performed another experiment with these
hyperparameters, where we learned the model $M$ with a new action space $A=\{ranker 2, ranker 3\}$ by
collecting user logs for two different non-overlapping time periods in which ranker 2 and ranker 3 were deployed. In this scenario, ranker 3 was more recently deployed relative to ranker 2. The test data is also generated
by collecting user logs on a hold-out time period where ranker 2 and ranker 3 were previously deployed. In this situation, the prior is that
the ranker 3 is on average better than ranker 2 and it has been incorporated in the
computation of the $L^1$-regularized and RE-regularized policies. The results are reported below in
Table \ref{tab:exp2}.


\begin{table}[h]
\centering
  \parbox{.48\linewidth}{
    \centering
    \begin{tabular}{l l}
      \hline
      \textbf{Algorithm} & \textbf{$\%$ improvement} \\ 
      & \textbf{ over ranker 3} \\ \hline
      unregularized policy &   $-0.1$     \\ \hline
      \textbf{$L^1$-regularized policy} &   $\textbf{0.214}$   \\ \hline
      \textbf{RE-regularized policy} & $\textbf{0.207}$ \\ \hline
      One-shot Policy &   $-6.33$   \\ \hline
      Regularized one-shot Policy &   $-2.92$   \\ \hline
    \end{tabular}
  }
  \caption{Performance comparison of different policies with pre-learned hyperparameters. The task
    is to identify the better search algorithm among ranker 2 and ranker 3 for a given query. The
    $L^1$-regularized and RE-regularized policies outperform other approaches as (a) they account for
    delayed rewards through MDP based session modeling and (b) they are robust to noise by factoring
    in the prior knowledge.}
  \label{tab:exp2}
\vspace{2mm}
\end{table}

We make several observations from the results listed in \Cref{tab:exp2}. First, there is a
significant difference between the performance of the regularized one-shot policy and the
regularized MDP policies, as the sessions were of longer range in the collected data. For example, a
typical query improved using older search algorithm is "mens gifts". As this query leads to sessions
of larger length on the shopping store, the regularized MDP policies allow us to factor in the
delayed rewards and suggest the appropriate search algorithm for the query. An empirical approach
such as $\text{OSP}_\lambda$ fails to evaluate the quality of search results in this case as it
focuses only on the immediate rewards. Second, the regularized MDP policies outperform both ranker 2
and ranker 3, whereas all other policies are worse of than the simple strategy of
selecting ranker 3 for all queries. This is because the $L^1$-regularized and
RE-regularized MDP policies account for the noise present in the model and incorporate the known
prior information appropriately.

\section{Conclusions}
In this paper, we study the problem of learning policies where the parameters of the underlying MDP
$\bar{M}$ are not known but instead estimated from empirical data. Simply learning policies on the
estimated model $M$ may lead to poor generalization performance on the underlying MDP $\bar{M}$. To
address this issue, we propose a Bayesian approach, which regularizes the objective function of the
MDP to learn policies that are robust to noise. Our learned policies are based on $L^1$ norm
regularization and relative entropic regularization on the objective function of MDP. We show that
our proposed regularized MDP approaches end up penalizing the reward of less preferred actions,
thereby giving preference to certain actions in $A$ over others.

To validate the performance of proposed algorithms, we evaluate the performance on both synthetic
examples and on the logs of real-world online shopping store data set. We demonstrate that a policy
learned optimally on $M$ without any regularization can even do worse than a simple policy that
always selects one of the actions $a \in A$ for all the states. Our experiments reveal that the
un-regularized policies are not robust to noise in probability and reward tensors. On the other
hand, the regularized MDP policies significantly outperform other baseline algorithms both on
synthetic and real-world numerical experiments.



\bibliography{mdp}

\begin{thebibliography}{}

\bibitem[Bell, 1965]{bell1965gershgorin}
Bell, H.~E. (1965).
\newblock Gershgorin's theorem and the zeros of polynomials.
\newblock {\em The American Mathematical Monthly}, 72(3):292--295.

\bibitem[Bellman, 1966]{bellman1966dynamic}
Bellman, R. (1966).
\newblock Dynamic programming.
\newblock {\em Science}, 153(3731):34--37.

\bibitem[Cooper and Rangarajan, 2012]{cooper2012performance}
Cooper, W.~L. and Rangarajan, B. (2012).
\newblock Performance guarantees for empirical markov decision processes with
  applications to multiperiod inventory models.
\newblock {\em Operations Research}, 60(5):1267--1281.

\bibitem[Dai et~al., 2018]{dai2018sbeed}
Dai, B., Shaw, A., Li, L., Xiao, L., He, N., Liu, Z., Chen, J., and Song, L.
  (2018).
\newblock Sbeed: Convergent reinforcement learning with nonlinear function
  approximation.
\newblock In {\em International Conference on Machine Learning}, pages
  1125--1134. PMLR.

\bibitem[Etter and Ying, 2021]{etter2021operator}
Etter, P. and Ying, L. (2021).
\newblock Operator augmentation for general noisy matrix systems.
\newblock {\em arXiv preprint arXiv:2104.11294}.

\bibitem[Etter and Ying, 2020]{etter2020operator}
Etter, P.~A. and Ying, L. (2020).
\newblock Operator augmentation for noisy elliptic systems.
\newblock {\em arXiv preprint arXiv:2010.09656}.

\bibitem[Fox et~al., 2015]{fox2015taming}
Fox, R., Pakman, A., and Tishby, N. (2015).
\newblock Taming the noise in reinforcement learning via soft updates.
\newblock {\em arXiv preprint arXiv:1512.08562}.

\bibitem[Ghavamzadeh et~al., 2016]{ghavamzadeh2016bayesian}
Ghavamzadeh, M., Mannor, S., Pineau, J., and Tamar, A. (2016).
\newblock Bayesian reinforcement learning: A survey.
\newblock {\em arXiv preprint arXiv:1609.04436}.

\bibitem[Gimelfarb et~al., 2018]{gimelfarb2018reinforcement}
Gimelfarb, M., Sanner, S., and Lee, C.-G. (2018).
\newblock Reinforcement learning with multiple experts: A bayesian model
  combination approach.
\newblock {\em Advances in Neural Information Processing Systems},
  31:9528--9538.

\bibitem[Grzes, 2017]{grzes2017reward}
Grzes, M. (2017).
\newblock Reward shaping in episodic reinforcement learning.

\bibitem[Haarnoja et~al., 2018]{haarnoja2018soft}
Haarnoja, T., Zhou, A., Abbeel, P., and Levine, S. (2018).
\newblock Soft actor-critic: Off-policy maximum entropy deep reinforcement
  learning with a stochastic actor.
\newblock In {\em International conference on machine learning}, pages
  1861--1870. PMLR.

\bibitem[Harutyunyan et~al., 2015]{harutyunyan2015shaping}
Harutyunyan, A., Brys, T., Vrancx, P., and Now{\'e}, A. (2015).
\newblock Shaping mario with human advice.
\newblock In {\em AAMAS}, pages 1913--1914.

\bibitem[Levine et~al., 2020]{levine2020offline}
Levine, S., Kumar, A., Tucker, G., and Fu, J. (2020).
\newblock Offline reinforcement learning: Tutorial, review, and perspectives on
  open problems.
\newblock {\em arXiv preprint arXiv:2005.01643}.

\bibitem[Mnih et~al., 2016]{mnih2016asynchronous}
Mnih, V., Badia, A.~P., Mirza, M., Graves, A., Lillicrap, T., Harley, T.,
  Silver, D., and Kavukcuoglu, K. (2016).
\newblock Asynchronous methods for deep reinforcement learning.
\newblock In {\em International conference on machine learning}, pages
  1928--1937. PMLR.

\bibitem[Nachum et~al., 2017]{nachum2017trust}
Nachum, O., Norouzi, M., Xu, K., and Schuurmans, D. (2017).
\newblock Trust-pcl: An off-policy trust region method for continuous control.
\newblock {\em arXiv preprint arXiv:1707.01891}.

\bibitem[Neu et~al., 2017]{neu2017unified}
Neu, G., Jonsson, A., and G{\'o}mez, V. (2017).
\newblock A unified view of entropy-regularized markov decision processes.
\newblock {\em arXiv preprint arXiv:1705.07798}.

\bibitem[Ng et~al., 1999]{ng1999policy}
Ng, A.~Y., Harada, D., and Russell, S. (1999).
\newblock Policy invariance under reward transformations: Theory and
  application to reward shaping.
\newblock In {\em Icml}, volume~99, pages 278--287.

\bibitem[Peters et~al., 2010]{peters2010relative}
Peters, J., Mulling, K., and Altun, Y. (2010).
\newblock Relative entropy policy search.
\newblock In {\em Twenty-Fourth AAAI Conference on Artificial Intelligence}.

\bibitem[Puterman, 2014]{puterman2014markov}
Puterman, M.~L. (2014).
\newblock {\em Markov decision processes: discrete stochastic dynamic
  programming}.
\newblock John Wiley \& Sons.

\bibitem[Schulman et~al., 2015]{schulman2015trust}
Schulman, J., Levine, S., Abbeel, P., Jordan, M., and Moritz, P. (2015).
\newblock Trust region policy optimization.
\newblock In {\em International conference on machine learning}, pages
  1889--1897. PMLR.

\bibitem[Sutton and Barto, 2018]{sutton2018reinforcement}
Sutton, R.~S. and Barto, A.~G. (2018).
\newblock {\em Reinforcement learning: An introduction}.
\newblock MIT press.

\bibitem[Wu et~al., 2019]{wu2019behavior}
Wu, Y., Tucker, G., and Nachum, O. (2019).
\newblock Behavior regularized offline reinforcement learning.
\newblock {\em arXiv preprint arXiv:1911.11361}.

\bibitem[Ye, 2011]{ye2011simplex}
Ye, Y. (2011).
\newblock The simplex and policy-iteration methods are strongly polynomial for
  the markov decision problem with a fixed discount rate.
\newblock {\em Mathematics of Operations Research}, 36(4):593--603.

\bibitem[Ying and Zhu, 2020]{ying2020note}
Ying, L. and Zhu, Y. (2020).
\newblock A note on optimization formulations of markov decision processes.
\newblock {\em arXiv preprint arXiv:2012.09417}.

\end{thebibliography}

\end{document}